# Types of Cognition and its Implications for future High-Level Cognitive Machines

**Camilo Miguel Signorelli**

Department of Experimental and Health Sciences, Universidad Pompeu Fabra, Barcelona, Spain
Centro Interdisciplinario de Neurociencias de Valparaíso, Universidad de Valparaíso, Chile
camiguel@uc.cl

**Abstract**

This work summarizes part of current knowledge on High-level Cognitive process and its relation with biological hardware. Thus, it is possible to identify some paradoxes which could impact the development of future technologies and artificial intelligence: we may make a High-level Cognitive Machine, sacrificing the principal attribute of a machine, its accuracy.

## Introduction

Intelligence is one of the main characteristics of the Human Brain. For example, cognitive abilities such as problem-solving, language skills, social skills, among others, are considered as High-level cognitive processes with a strong impact on different types of intelligences. Understanding how these abilities work can help us to implement desirable social behaviors in the new generation of robots with Human-Machine interactions.

How do cognitive, abstract abilities and subjective experience emerge from a physical system like the brain? (Chalmers, 1995; Dehaene, et al. 2014; Tononi, et al. 2016) How can we implement these abilities in a natural form in machines? Is it possible? What are the requirements to achieve this? These are some of the most complex questions in science and to resolve them, we need to understand the biological basis, the cellular dynamics as well as the physical and mathematics basis to know the fundamental properties of matter which could eventually be involved in these emergent properties.

One requirement for High-level cognitive abilities seems to be what it is called subjective experience (Barron & Klein, 2016) or in a more complex order: consciousness. For example, we first need to be conscious to take some complex rational decisions, to plan, and to do something with intent (Baars, 2005; Tononi & Koch, 2008). How biology implements rational or high intelligences, is completely different from how we started to implement it in computer science (Moravec, 1988).

In order to improve machines socials skills, one should try to understand how these abilities emerge in the brain and also, in order to understand how these abilities work and emerge from the brain, we should study how to implement them into machines. This is why the last test or full understanding of subjective experience and consciousness will be to make a conscious machine.

This work will show a brief of what it is known about conscious states and some paradoxes that emerge from the idea of creating a conscious machine: it may be possible to make a conscious robot, but it does not come without a cost.

## Consciousness and Intelligence

**Consciousness, a brief notion**

If consciousness could be defined, the questions above would be easily answered. We may also have a recipe with different "ingredients" to explain consciousness in humans, animals, and accept or reject the option to implement it in machines.

However, this is not the case. All of us understand what consciousness is, what it is to be aware, what it is to have a subjective experience and on which "elements" our experience is based. Sometimes these concepts are considered as synonyms, other times as different ideas with different meanings. Thus, it is needed to make some distinctions. The modern approach to consciousness has made a differentiation between **conscious states** as different levels of awareness (vegetative, sleep, anesthesia, altered states, aware); **contents of consciousness** as elements or information in the external or internal world which at times are part of our conscious state; and **conscious processing** as

---

the operations applied to these contents (Dehaene, et al. 2014). Thanks to these definitions, theories as Global Workspace (Baars, 2005), Integrated Information (Tononi, et al. 2016) and Dynamic theories (Varela, et al. 2001) have developed many tools leading to a greater understanding of consciousness. In these approaches, the consciousness and eventually intentionality are understood as an emergent property from brain networks dynamics.

Nevertheless, these definitions have some difficulties explaining how this emergency works, explaining subjectivity experiences (Chalmers, 1995; Lycan & Dennett, 1993) or sometimes avoiding the importance of previous learning to develop consciousness at early stage of childhood (Cleeremans, 2011).

Even though this article will refer to the above definitions, as we are referring to previous works that use these definitions; our own understanding of consciousness, unlike these theories, might be defined as a process of processes which mainly interferes with the neural integration. In this case, consciousness would be an operation more than a container where one might put something. It implies that consciousness could emerge from neural network dynamics as an autonomous system (e.g. brain oscillations), at a different space of action than the original system, but with sub-emergent properties interfering with the space from which it originates. This work will interpret certain results in this direction with the intention of expanding this notion in future projects. Also, while the author agrees with many of the conclusions, results and progresses of the theories mentioned, it is reasonable to consider that a new focus that integrates all of them is needed, since they share many similarities and are at the same time complementary.

The concept of consciousness involves awareness, emotions, subjectivity, intentionality, and attention, among others. Consciousness is formed from all of these processes like a differentiated and unified whole, but it is not any of them. For example, it could be necessary to be aware to have emotions and subjective experiences, or maybe vice versa, and we will need them to show intentionality and attention. We also distinguish that these are different processes as, for example, awareness and attention; while understanding all of them as constituent parts of what we define as consciousness.

**Consciousness and its relation with Intelligence**
Intelligence is something a little simpler to define; at least, many scientists think so. Intelligence can be understood as the ability to solve problems in an efficient way. This means, the maximization of the positive results in our solution while minimizing the negative impacts, for instance, waste of time. To do that, other processes as Learning and Memory are also needed and associated with the definition of intelligence. Learning may be understood as the process to gain new knowledge or improve some behavior, while Memory is the storage of this knowledge. To solve problems efficiently, it is necessary to access a certain memory that was acquired thanks to a specific learning that will be modifying again the memory of the system. The more intelligent is the system, the more it learns.

This framework also allows finding mathematical implementations of intelligence and its use in machines. Why could consciousness be necessary in this context? First, many different types of intelligences have already recognized, between them, the emotional intelligence that helps us with social interaction. Secondly, emotions play a crucial role in learning, consolidation of memories, retrieved memory and cognition in general (Cleeremans, 2011). In other words, emotions, as subjective experiences, and cognition are deeply related into human intelligence (Haladjian & Montemayor, 2016).

Finally, emotions and subjective experiences are also constitutive parts of consciousness, as stated above. This means consciousness and High-level cognitive abilities are also related. In consequence, to improve complex intelligence and social behavior in machines, we should understand constitutive parts of consciousness, and how its processes affect the computability and accuracy of the most powerful system that we know to date, our brains.

## What we already know
### First: The Hardware
One simplification about the brain code is to think in the neuron as the principal informative unit for the brain codification and the spikes as the language of its code. It assumes one kind of codification between 0, without spike, and 1 with a spike.

However, now it is known that the electrical transmission through gap junctions (Bennett & Zukin, 2004) are not only between other neurons but also between cells as astrocytes (Alvarez-maubecin, et al. 2000). It is also known that neuromodulatory substances can reconfigure neuronal circuits within minutes and hours (Nusbaum, et al. 2001). In some cases, these substances applied in axon (Bucher, Thirumalai, & Marder, 2003; Goaillard, et al. 2004) can spontaneously initiate action potentials in a non-classical mode of integration. Another research has shown that action potential in some neurons can travel backward from the axon and soma regions into dendrites (Stuart, Dodt, & Sakmann, 1993), in contrary to Polarization law. Other examples are: The functional complexity of dendrites and the roles they play, the influence of non-neuronal cells in the action potential propagation (Dupree, et al. 2005) and axon-glia communication, where the information is transduced through cells that are not neurons (Fields & Stevens-graham, 2005). Astrocytes synapses regulation (Coggan, et al. 2005; Dani, Chernjavsky, & Smith, 1992; Fields & Stevens-graham, 2005; Matsui & Jahr, 2004) and its influ-ence on neurons and so on. All this new knowledge is not included in the idea of spike codification or is definitely contrary to it.

These findings have shown us that the neuron is not the unique codification unit in the brain code and in consequence the spike is not the unique key to understand its codification (Llinas, et al. 1998). It implies that the binary code is not enough to understand the brain.

**Second: Computability**
Turing and Gödel were the first in demonstrating the theorem of incompleteness (Gödel, 1931; Turing, 1937). This theorem showed that it is not possible to resolve all the problems with a deductive process or algorithm. From their demonstrations, one of the consequences suggested is the potential impossibility to build conscious machines with algorithms only. Physiological and Psychological data also seem to show that this is not possible (Haladjian & Montemayor, 2016). This, together with point one, means that if we can reach the gap to build a conscious machine, it will not be with the algorithmic computation. A new computational framework will be needed.

**Third: The two Systems**
When we try to explain, on one side, the biological mechanisms in the brain, and on the other, the human psychological behavioral, some paradoxes appear. Some research and theories have shown that the dynamic of neural systems can be interpreted in a classic probabilities framework (Pouget, et al. 2000; Quiroga & Panzeri, 2009), like good estimator and predictor of external stimuli. While other results, principally from economic psychology, show cognitive fallacies (Gilovich, et al. 2002, Ellsberg, 1961; Machina, 2009; Moore, 2002). These results are incompatible with the classical probability theories (Pothos & Busemeyer, 2013).

Hence, these disconnections between the behavior of the human brain and the cognitive fallacies may help to understand better how the human brain works. How can some cognitive capabilities, with apparently non-classical dynamic, emerge from apparently classical systems as neural networks? It is necessary to answer this question as a requirement to start to think in conscious machines.

**Forth: Four types of cognition**
A recent paper by Shea and Frith (Shea & Frith, 2016) presents a valuable differentiation between contents and cognitive processes (computations from some contents to others). These categories can be conscious or not. From the experimental evidence and with this framework, they define four cognitive categories (Table 1): i) Type 0 Cognition: non-conscious contents and non-conscious (non-controlled) processing. Optimal choice is computation-light but learning-heavy (e.g. Motor Control). ii) Type 1 Cognition: conscious contents but automatic and non-controlled processing (e.g. Fallacy questions). The system accesses to a wider range of information (Holistic information), however some optimal calculations may become computationally intractable. iii) Type 2 Cognition: contents are conscious and the cognitive process is deliberate and controlled (e.g. Reasoning). It is computation-heavy, learning-light and interferes with Type 0. iv) Type ∞: non-conscious content but conscious process. Yet no further definition is given as they are not sure if this category could exist (Supra Reasoning information).

|  | Non conscious Processing | Conscious Processing |
|---|---|---|
| Contents non conscious | Type 0 Cognition | Type ∞ Cognition |
| Contents conscious | Type 1 Cognition | Type 2 Cognition |

**Table 1.** Types of cognition according different contents and processing exposed above.

**Fifth: The perception is a discrete process**
The perception needs time. We are conscious of a perceptual content in more o less 200 ms up 400 ms. Indeed, evidence suggests that the conscious perception is a discrete process (Dehaene & Changeux, 2011; Dehaene, et al. 2014; VanRullen & Koch, 2003) in contrast to a continuous process. The differences between fast time processing for cognition type 0 (order of 40 ms) and a slow time processing for type 1 (order of 200 ms) have stimulated the idea of the Two-Stage Model (Herzog, Kammer, & Scharnowski, 2016). The first stage would be a fast, continuous and classic probabilistic process (type 0 cognition) while the second stage would correspond to a slow and discrete process (type 1 and 2 cognition) due to loops in information processing (Figure 1). In relation with the discussion above, one may also say that the second stage would obey non-classical probabilities in some context.

## Implications and Paradoxes

The four types of cognitive categories, together with the other points that were mentioned previously, open the door for some new interpretations and classifications on the types of machines (Figure 1):

- The Machine-Machine: It could correspond to the type 0 cognition. The examples are the robots that we are making today with a high learning curve.
- The Conscious Machine: A machine with the ability to be conscious of its contents but not of the process. In this case, we will sacrifice accuracy in favor of consciousness. The type 1 cognition requires integration of information and it is a time consuming process. This is why the accuracy is lost instead of consciousness. However, is this machine useful for something? Here is the first Paradox. If we can make a Conscious Machine Type 1 Cognition, this machine will lose the most important attribute of being a machine: the accuracy in the calculation. In this case, the machine does not work how we expect: it is a machine without being a machine.

- The Super Machine: A machine with the ability to be conscious of its contents and own processes. This machine will have subjective experience at the same time that it will have the option to control the accuracy of its own rational process. However, the type 2 cognition needs more time to find accuracy solutions than type 1. This process also interferes with type 0 cognition, and in consequence robots will have less precision in motor tasks. In this case, one observation appears: we cannot expect an algorithm to control the process of emergency of consciousness (point 2) and we will not be able to control it. In other words, the machine can do whatever it wants; it has the power to do it and the intention to do it. It could be considered biological new species, more than a machine. The paradox is that the Super Machine is not useful anymore unless the machine wants to collaborate with us.
- The Subjective Machine: Machine type ∞ cognition. Some modern progress in artificial intelligence might fit in this category (Bringsjord, et al. 2015). The machine will be conscious of each computation process but not conscious of the contents, at least, they will not be aware in the sense that we are. What does this mean? Probably the Machine will be able to control the computational processing; however it will not be able to take meanings of contents from these computations. In other words, the machine will not be able to "feel something", to have "emotions", the machine will have a type of subjective experience that we do not know yet. This is probably the evolution of machine type 0 that we are making these days but without the intermediate steps type 1 and 2 cognitions. However, if this machine does not feel like a human, will it be able to understand us? It could be the kind of machine in science fiction. Again, the machine could be useless for our purpose and even worst, it could be contrary to our purposes.

Based on these observations, it is possible to summarize all the paradoxes in one type: The Conscious Machine is not a useful machine anymore. In other words, the machine losses the meaningful characteristics of being a machine: to resolve problems with accuracy, speed and obedience.

## Some Alternatives

It could be interesting for Artificial Intelligence to make Machine type 1 or type 2 cognition. Based on the principles exposed above from biological systems, the requirements are: i) new non-binary codification principles, ii) no deterministic computation, iii) two different systems of integration of information, which should be related as the non-classical probabilistic system emerging from the classical probabilistic system, iv) enough computation capability to support the first time-consuming system and the second non-classical system. Is it possible to respond to these requirements with the actual framework in Artificial Intelligence?

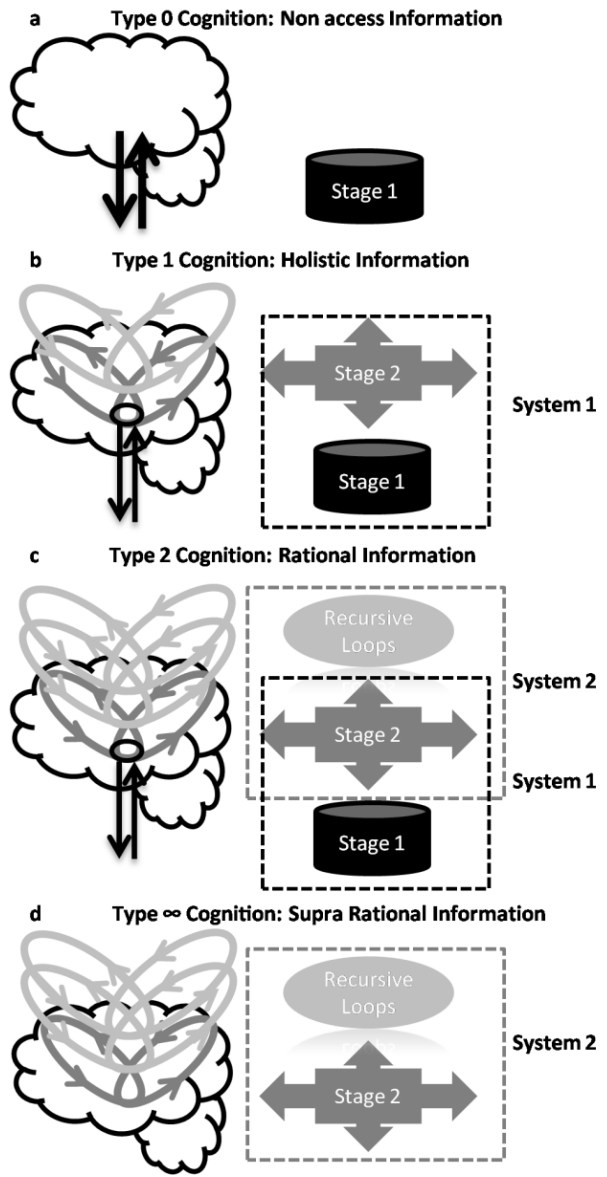

**Figure 1.** Types of cognition related with potential Systems of processing in the brain.

Quantum computation is one of the alternatives due to the increase of computational capability thanks to qubits, non-classical probabilities and non-classical computability. Some alternative tools aiming to implement Quantum computation in analogy with brain process, might be new approaches of contextuality from Quantum Physics to Psychology (Abramsky, 2015), like contextual semantics, natural language, and semantics of computation (Kartsaklis, et al. 2013). Starting from the explanation of the cognitive fallacies above; concept combinations, human judgments and decision making under uncertainty can be better modeled and described by using the mathematical formalism of quantum theory (Aerts, et al. 2013; Busemeyer, et al. 2011; Sozzo, 2015; Wang, et al. 2014). This emerging domain has been called "quantum cognition" (Bruza, Wang, & Busemeyer, 2015) and these models may help us understand the connection between new computational frameworks, psychophysics, recent advances in neuroscience on conscious perception in the brain, and also how to make conscious machines.

## Conclusions

These ideas are part of a work in progress. This work has developed a hypothesis based on previous works: we may build conscious machines but with computational costs. These costs reflect some paradoxes when we examine the requirements to build a conscious machine: it will be possible to make a conscious machine without good rational abilities or a good rational conscious machine with more time to respond accuracy. In both cases, the price of non-obedience will be paid. The next step in this research will be to quantify the decrease in computational efficiency, time-consuming of these processes and the mathematical formulation of these propositions.

## Acknowledgments

This work was partially supported by CONICYT. The author appreciates valuable comments of Marlène Morvan and Miriam Potter in the final version.